# Reward Systems for Trustworthy Medical Federated Learning

Konstantin D. Pandl, Florian Leiser, Scott Thiebes, and Ali Sunyaev

Karlsruhe Institute of Technology, Germany

Federated learning (FL) is a popular technique to train machine learning (ML) models for healthcare where institutions collaborate by sharing ML model gradients. An undesired phenomenon in ML models is bias, which may cause unfairness against specific subgroups. In FL research, it remains unclear to which extent bias occurs, and how to best incentivize institutions contributing toward trustworthy FL models. Besides bias, trustworthiness aspects include high predictive performance. Existing reward systems, incentivizing predictive performance only, can transfer model bias against patients to an institutional level. Therefore, we evaluate how to measure the contributions of institutions toward either predictive performance or bias in FL and design corresponding reward systems, before we propose a combined reward system that incentivizes both. We evaluate our work using multiple chest X-ray datasets focusing on sex- and age-related patient subgroups. Our results show that we can successfully measure contributions toward bias, and an integrated reward system successfully incentivizes contributions toward a well-performing model with low bias. Thereby, institutions with data predominantly from one subgroup introduce a favorable bias for this subgroup. In a label flip experiment, we furthermore show that reward systems can incentivize institutions to ensure a high dataset quality.

**Keywords:** Bias, Federated Learning, Incentives, Medical Imaging, Reward Systems

## 1  INTRODUCTION

Federated learning (FL) is a widely debated technique for machine learning (ML) in healthcare, as it enables the collaborative training of ML models. In doing so, FL can help break up data silos defined by institutional boundaries, allowing for ML models to be trained on more data compared to single institutions [1, 2]. In theory, the increased amount of training data should help to improve the performance of ML models. Moreover, this larger data collection is potentially more diverse and can potentially help with reducing bias in ML models (i.e., disparities in the predictive performance of the ML model across different (sub-)groups) [1]. In FL practice, data across different FL clients are typically not independent and identically distributed due to different patient demographics at institutions, different data acquisition devices (i.e., scanners in the case of medical imaging), or different data labelling practices. This generally leads to a degradation of the predictive performance of federated models. Still, such federated models potentially perform better than ML models trained on data from a single institution only [1, 3]. However, a high predictive performance is not the only important aspect to achieve trustworthy AI in healthcare [4]. Ensuring a low model bias is of paramount importance to treat different patient subgroups fairly and to obtain their trust. So far, the impact of FL on model bias is little studied. On the one hand, there are hopes that the more diverse training data can help in reducing bias [1]. On the other hand, there are concerns that FL can potentially amplify bias of ML models. For example, ML models may perform well for most institutions but poorly for institutions with minority patient demographics leading to a worsened performance of such ML models at these institutions, with potentially severe patient consequences [5, 6]. Reducing prediction bias is therefore important and in line with an increasing emphasis on trustworthy AI [4], and to enable the adoption and harvesting the benefits of ML-based systems in healthcare practice [7].

A possible solution to reduce the bias of ML models in FL lies in the introduction of reward systems to better incentivize the desired contributions of institutions participating in FL [4]. Generally, active participation in FL comes with costs and risks to institutions (e.g., data collection, compute costs, privacy risks) [1, 8]. The individual contribution benefits for institutions are small, which can lead to a lack of willingness to participate actively in FL. This is also known as the free-rider problem [9], which also occurs in medical data-sharing scenarios [10]. Considering this, reward systems can reimburse institutions for their efforts and risks, and promote active participation in FL.

Extant works on reward systems in FL networks usually focused on the institutions' contributions to predictive performance [11-13]. Reward systems only incentivizing a high predictive performance can lead to FL models performing well on most test data, but performing poorly on test data of specific subgroups, creating bias and potentially resulting in severe health consequences for these subgroups in practice. Therefore, incentives of a high predictive performance and a low model bias must be well-aligned to obtain desired outcomes. As incentives rely on contribution quantification, we argue quantification of institutions' contributions to lower model bias should also be represented in reward systems.

Extant research uses the Shapley Value (SV) to estimate the contributions of each institution which uniquely fulfills desirable properties of collaboration fairness [9]. However, computing SVs is computationally complex, therefore, we rely on previous efforts to develop efficient approximations [11, 14, 15]. These approximations have focused only on contributions toward predictive performance so far. It remains unclear how contributions toward bias in an FL scenario could be quantified, which we investigate in a first step. Subsequently, it remains unclear how to translate the contributions into rewards for institutions. Thus, we develop different reward systems that enable the compensation of institutions. Such a reward system can incentivize the result of well-performing ML models in FL not only in the sense of a high predictive performance, but also simultaneously a low model bias. As such, it can ultimately lead to more trustworthy ML models.

To conduct our research, we train a convolutional neural network using FL on three chest X-ray datasets and quantify predictive contributions using the SV approximations. We then translate these quantifications into a reward (e.g., monetary units). After that, we analyze contributions toward the bias of the FL model with SVs concerning sex-based and age-based subgroups. Then, we propose a reward system that translates SVs into rewards and incentivizes both, a high predictive performance, and a small absolute model bias. Finally, we analyze the reward systems in the experimental settings, including in a label flip experiment where we analyze incentives with regard to dataset quality.

In doing so, we make three key contributions to the extant body of knowledge on FL. First, we contribute to a better understanding of the occurrence and extent of model bias in medical FL scenarios, answering prior calls for research in this area [1, 5, 6]. Second, we show that SV approximations can also quantify contributions toward bias in medical FL. Third, we develop an incentive scheme that rewards predictive performance contributions and model bias contributions. Thereby, we show practical applications and we answer previous calls for research on reward systems for FL [1, 9, 16] and incentives for trustworthy AI [4].

We organize the remainder of this paper as follows. In section 2, we provide a background on bias in medical AI, FL, contribution quantification for FL, and reward systems for FL. Afterward, we describe our methods in section 3. In section 4, we describe the results of the contribution quantification. This is followed by section 5, where we develop reward schemes based on contribution quantification, and evaluate their utility on real-world medical data. In section 6, we discuss our results, whereas section 7 concludes the paper.

## 2 BACKGROUND

### 2.1 Bias of machine learning models in medicine

Bias of an ML model refers to a disparate predictive performance across different groups [17]. A group can be defined by features such as sex, age, ethnicity, or economic welfare, among others. Different metrics can be used when evaluating bias, for example, accuracy, the area under the receiver operating characteristic (AUROC), or sensitivity [17]. A small bias is often desired to avoid discrimination of certain groups, ensure fairness, and ultimately, enable trustworthy ML. Bias is critical in healthcare, as the deployment of ML can affect the health of people [7].

Prior research has identified various biases in ML algorithms for medical image classification. For example, chest X-ray classifiers trained on multiple datasets had an unfavorable sensitivity for female patients, Hispanic patients, and patients with Medicaid insurance [18]. This research also showed that acquiring data from multiple sources can reduce this bias [18]. Other research discovered that dataset imbalances can lead to the training of biased ML classifiers [19], suggesting that balanced datasets are essential to reduce bias.

The influence of FL on bias is an open field of research. On the one hand, there are hopes that more diverse training datasets from different FL institutions can reduce bias [1]. On the other hand, there are concerns that data heterogeneity among clients in FL might introduce additional bias [1, 6].

### 2.2 Federated learning

FL enables the training of ML models across data silos from different clients. The clients collaborate by training ML models locally on their data, and by sharing the gradients of the ML model parameters with a central server. This central server then averages the parameters across models from different clients and provides the resulting model back to the clients. This process is referred to as a communication round. To train an FL model, multiple communication rounds are performed.

With its ability to train well-performing ML models while preserving patient's privacy, FL is especially promising for healthcare applications [1]. In healthcare, many FL scenarios can be described as cross-silo FL [5, 20]. In contrast to cross-device FL, cross-silo FL is characterized by a low number of clients (i.e., 2-100), and a high client and network reliability [5]. In this work, we also refer to FL clients as institutions [5].

### 2.3 Contribution quantification for federated learning

When quantifying the contributions of FL participants to distribute rewards, collaborative fairness is important, meaning the quantified contribution should fairly reflect the quality of the actual gradient contributions of the FL participants [9]. Otherwise, institutions in a FL consortium may not agree on a procedure to quantify contributions. Research predominantly uses the SV because of its desirable properties [9]. In the present context, the SV for an institution $z_i$ is defined as the average marginal contribution of that institution to all possible subsets S of the set of all institutions D. This is also specified in the following formula, where N is the total number of institutions, and U is a performance metric [14, 15].

$$\varphi_{shap}(z_i) = \frac{1}{N} \sum_{S \subseteq D \setminus \{z_i\}} \frac{1}{\binom{N-1}{|S|}} [U(S \cup \{z_i\}) - U(S)] \tag{1}$$

Following this definition, the SV uniquely satisfies desirable properties, two of which are especially relevant in the context of reward systems. First, group rationality, meaning the entire value gain of the FL consortium, is completely distributed among all participating institutions (2).

$$U(D) = \sum_{z_i \in D} \varphi(z_i) \tag{2}$$

Second, fairness, meaning the value of an institution $z_i$ should be zero if the institution has zero marginal contribution to all subsets of the consortium: $\varphi(z_i) = 0$ if $U(S \cup \{z_i\}) - U(S) = 0$, for all $S \subseteq D \setminus \{z_i\}$. At the same time, the contribution quantification of two institutions $z_i$ and $z_j$ should have the same value if they both add the same performance to a subset of the consortium: meaning, $\varphi(z_i) = \varphi(z_j)$ if $U(S \cup \{z_i\}) = U(S \cup \{z_j\})$, for all $S \subseteq D \setminus \{z_i, z_j\}$ [14, 15].

However, the computation of SVs requires the training and testing of $2^N$ FL coalitions of institutions, which presents a very high computational and network burden for practical use cases. Therefore, efficient approximation of SVs is an actively evolving field of research with many recent, new approximation methods [11, 13, 21].

Pioneer work in this direction proposed two approximations [11]. Thereby, only the largest FL coalition is trained. Approximated FL models from all other coalitions are then computed by adding the gradients of the participating institutions from the largest coalition, over all communication rounds from training the largest coalition. The performance metric of the resulting models is then obtained from the test datasets. Based on the results, the SVs can be computed. As such, it requires the training of only one coalition, which is computationally and communication-wise the most expensive task in FL. Still, $2^N$ FL institutions need to be tested which usually also requires some computational effort. As such, recent work aims to further reduce the number of testings [13].

A novel method aims to further speed up the process by massively reducing the time required for individual testings and develops a technique named SaFE for the SV estimation specifically for institutional healthcare settings [21]. Thereby, the federated model is trained normally by the largest coalition. Afterward, the institutions extract deep features of the training data and testing data using the federated neural network (i.e., the activations of the last layer or a layer close to the output of the network given the respective input data). Then, a logistic regression (LR) model is trained over the deep features. The individual LR models are then exchanged to the server and tested. To obtain the utility of arbitrary coalitions, the LR outputs of the individual clients can be accumulated. This accumulation process is very quick compared to the actual testing of the federated model on the actual data. As such, this approach scales well for FL healthcare use cases, despite still having exponential complexity.

### 2.4 Reward systems for federated learning

Reward systems aim to incentivize contributions based on a performance/achievement-related pay principle, meaning participants get rewarded for their cooperation in accordance with their contributions. Application examples include mobile crowdsensing [8], distributed energy storage systems [22], or data marketplaces for centralized machine learning [23, 24].

In FL, research interest in reward systems is growing [9], in particular for healthcare [1]. Rewards can take different forms, for example, monetary value [1, 9], or reputation [25]. The fair distribution of rewards depends on a fair quantification of the contributions. The general goal of reward systems in FL is to compensate institutions for the effort in participating in FL [1], as well as to attract high-quality gradient contributions by institutions [8].

Beyond these factors, further unique challenges in FL demand for reward systems. Most contemporary FL systems give all participants access to the model, independent of their gradient contributions. Thus, these FL systems lack incentives for institutions to participate actively in the learning process. In view of potential privacy risks when publishing gradients in FL [26], there can be even benefits to not contribute actively to FL. This is also referred to as the free-rider problem [9], which is relevant in medical data-sharing scenarios [10]. Research on how to design reward systems for FL based on contribution quantification is in its early stages. Liu et al. [27] develop a blockchain consensus mechanism that incorporates SV computations for FL. Multiple rewards are thereby distributed. This includes a price paid by an FL task requester, where the price is then distributed proportionally based on the FL SVs for all clients with positive SVs. An FL task requester pays a price, which is then entirely distributed among the FL clients with a positive SV. Thereby, the rewards are proportional to an FL client's SV. Furthermore, a recent article by Zhang et al. [25] develop a reward system where FL clients bid their prices, and a selector selects clients to participate in an FL round. The reputation of clients and the payment of rewards depend on the quality of the contributions, which is evaluated by a ranking of the cosine similarity of client's gradients and the final FL model. In Gao et al. [28], the authors evaluate different contribution evaluation mechanisms toward model accuracy with an emphasis on detecting attacks by malicious institutions and excluding malicious institutions from the training process. Rewards are paid proportionally based on the reputation and contribution of institutions.

## 3 METHOD

### 3.1 Datasets

Different medical institutions typically have a heterogeneous data distribution, due to different medical scanning devices, patient demographics, or labeling practices [1]. To conduct realistic FL experiments, we use three large chest X-ray datasets, namely the NIH ChestX-ray8 (NIH) [29], CheXpert (CXP) [30], and MIMIC-CXR (CXR) [31]. These datasets contain chest X-ray scans, corresponding labels with information about the patient (e.g., patient ID, sex, age), and the medical condition visible in the scan. An overview of the dataset sizes and their distributions is provided in Table 1. Since the datasets have slightly different medical condition labels, our research focuses on the overlapping set of eight labels. This includes a *no finding* label, and seven disease labels for *atelectasis*, *cardiomegaly*, *consolidation*, *edema*, *pleural effusion*, *pneumonia*, and *pneumothorax*. The label values can be 1 (clinical observation present in the scan), 0 (clinical observation not present in the scan), -1 (no clear indication of the presence or absence of the clinical observation in the scan), or NaN (no information available) [29-31]. To obtain labels for our learning task, we follow prior research and transformed all non-1 labels to 0 [18, 19, 30].

Table 1: Overview of the three datasets used for our experiments.

| Metric | NIH [29] | CXP [30] | CXR [31] |
|---|---|---|---|
| Number of scans | 112,120 | 223,414 | 377,095 |
| Share of scans with sex specification [%] | 100 | 100 | 92.2 |
| Of these, number of scans from female / male patients | 48,780 / 63,340 | 90,778 / 132,636 | 164,817 / 182,915 |
| Share of scans with patient age information [%] | 100 | 100 | 99.8 |
| 30% youngest quantile, age [years] | 38 | 52 | 52 |
| 30% oldest quantile age [years] | 57 | 71 | 71 |

### 3.2 Data splits

We use 20% of each of the three datasets as a test dataset and the remaining 80% for the training and validation dataset. We split all datasets on a per-patient basis to avoid multiple scans from the same patient in different data splits.

In the first set of experiments, we split the training and validation sets based on patient sex (female and male). In the second set of experiments, we split it based on patient age (30% youngest patient scans and 30% oldest patient scans). In both scenarios, we obtain six FL institutions, which is a typical consortium size for cross-silo medical FL projects [5, 20].

We obtain two institutions participating in FL from each dataset. For the NIH dataset, for example, we refer to these as NIH-1 and NIH-2. We then use four different splits based on patient sex. First, an 'as is' split, where we use the original distribution for both institutions of each dataset (i.e., 43.5% female patient scans and 56.5% male patient scans for both NIH institutions). Second, a 50/50 split, where 50% of each institution's scans are from one class (e.g., female patients), and 50% are from the other class (e.g., male patients). Third, a 75/25 split, where one institution has 75% of the scans from class 1 (e.g., female patients) and 25% from class 2 (e.g., male patients). The other institution then has 25% of the scans from class 1 (e.g., female patients) and 75% of the scans from class 2 (e.g., male patients) of the same dataset. With this split, we can evaluate whether an increased data imbalance for the two institutions influences their bias contributions while maintaining the data distribution in the network. Fourth, we use a 100/0 split, where one institution contains scans from only one class (e.g., female patients), and the other institution contains only scans from the other class (e.g., male patients). We repeat the same procedure of data splits based on patient age (young and old patient subgroups) instead of sex. Therefore, we obtain eight different data splits in total, four based on patient sex, and four based on patient age.

Within these data splits, the size of the training and validation dataset is defined by the smallest subgroup. For the sex-based splits, this is the set of female patient scans of the NIH dataset (i.e., 38,776 scans after accounting for a 20% test dataset, cf. Table 1). To account for differences when splitting the dataset per patient, we use 35,000 scans per institution, 80% (28,000) for the training dataset, and 20% (7,000) for the validation set. For age-based splits, we split the data based on quantiles rather than fixed ages, given the different age distributions per dataset. The NIH dataset is the smallest institution dataset regarding patient age with a maximum of 26,908 scans for the training and validation sets. Again, we used an 80%/20% split with 20,800 scans for the training, and 5,200 scans for testing.

We use the same dataset size for all institutions. This allows us to distill the influence of data partitioning on the bias, and to avoid an influence of different dataset size distributions, which can further influence the bias in a federated weighted averaging aggregation scenario [6].

### 3.3 Experimental setup

For our ML task setup, we use a DenseNet-121 architecture [32] pre-trained on the ImageNet dataset, as it has demonstrated effectiveness in chest X-ray classification tasks [18, 33]. To preprocess scans for the training task, we follow prior research [18] and use a random horizontal flip and a random orientation of up to 15°. Furthermore, we use a random shift of up to 10% of the image height and width and a batch size of 32 images. We also scale all dataset images to 256x256 pixels and apply normalization based on the ImageNet dataset for all datasets.

For FL, we use the federated averaging algorithm [34] in a simulation on a single computing machine. In our implementation, each institution possesses a large training and a smaller validation dataset. The institutions train the ML model locally for one epoch with a stochastic gradient descent optimization algorithm at a learning rate of 0.1 and a binary cross entropy loss function. After each communication round, every institution evaluates the newly aggregated global model on its validation dataset. The federated training stops when there is no improvement in the average validation loss for the last 10 communication rounds.

In the testing procedure, the institutions use the global model with the lowest validation loss, averaged across all institutions. We evaluate our tests using the AUROC metric on the entire test dataset and on subsets based on sex and age. The AUROC generally ranges from 0 to 1, with a random classifier exhibiting a value of 0.5. A higher value indicates a better predictive performance. We use the AUROC metric for three reasons. First, it is expressive even for imbalanced datasets, which is often the case for medical imaging datasets. Second, it condenses the information of the receiver operating curve into a single, lucid number. Third, it is widely used for AI in medical imaging [33] and bias in medical image AI [19]. This enables researchers to compare and classify our results with existing research more effectively.

We quantify the bias in the data splits with varying patient sex distribution as the difference in the AUROC between the two subgroups. In our case, we measure the AUROC of female patient test scans minus the AUROC of male patient test scans. Thus, a positive-valued bias indicates a better predictive performance of the classifier on female patient scans. In contrast, a negative-valued bias indicates a better predictive performance on male patient scans. A bias of zero indicates no difference in the predictive performance on male or female patient data. Accordingly, for varying patient age distributions, we define the bias as the AUROC of the classifier on the young patient test dataset minus the AUROC of the classifier on the old patient test dataset.

To compute SVs efficiently, we follow the algorithm described in section 2.3 [21] for three reasons. First, it has shown to provide accurate SV estimations [21]. Second, it is computationally efficient [21]. Third, potential shortcomings such as a lack of evaluation in each communication round do not negatively impact our case of quantifying contributions in a healthcare scenario as cross-silo medical FL is often characterized by reliable institutions [5]. In the implementation, we compute approximated LR model outputs of all 63 possible coalitions of the six institutions. We then test these regarding general predictive performance and bias, and compute SVs for both.

We repeat our experiments 40 times with different random seeds, to compute 95% confidence intervals for our results. To analyze the statistical significance of certain observations, we conduct paired t-tests, significance thereby occurring at a p-value of less than 0.05.

Based on the computations, we develop reward systems to incentivize a high predictive performance in FL, a low bias, and both in combination. To show how such incentives are effective in healthcare practice, we also conduct a label flip experiment. In healthcare practice, datasets often have erroneous labels, with one study estimating ca. 7.4 % wrong labels in the CXP dataset [35]. These erroneous labels generally lead to lower ML model performance and robustness. Thus, ensuring a high label quality in ML training datasets is important to obtain well-performing, robust, and ultimately trustworthy ML models. However, it requires medical experts to analyze and re-assess data. As such, it is expensive and binds scarce resources in healthcare institutions. Despite different heuristics from research to improve efficiency in this process by identifying likely mislabeled data points [15, 35], it can remain a cumbersome process [36]. As label errors in a single institution likely have only a small negative influence in FL, institutions have little incentive to ensure they use error-free datasets, which can become a problem. As such, reward systems could help institutions, if the rewards for predictive performance contributions for datasets with high label quality are higher than for datasets with low label quality. To evaluate this, we assume valid labels in each dataset and conduct experiments, where three clients have unchanged labels, and three counterpart clients sampled from the same datasets have erroneous labels introduced with a certain ratio of flipped labels. Specifically, we flip labels in a range of 2.5 %, 5 %, and 7.5 % which is a realistic range for the frequency of label errors [35]. We then compare the received rewards of the flipped clients with the unchanged clients. We conduct the label flip experiment in each setting 12 times to obtain means and confidence intervals.

During the experiments, we also measure the execution time of the data valuation approach to examine the scalability of our approach. We publish the source code of our experiments, including the random seeds of the data splits, on GitHub[1] - this allows future research to reproduce and build on our results. We run our work on a computing cluster consisting of NVIDIA Tesla V100 and NVIDIA GeForce RTX 3090 GPUs using Python 3.7.11 and PyTorch 1.10.2.

## 4 CONTRIBUTION QUANTIFICATION RESULTS

### 4.1 Contributions toward predictive performance

The results for the contribution quantification regarding predictive performance, meaning the SVs, are shown in Table 2. In all cases, the mean total AUROC is higher for classifiers trained on sex-based splits, where institutions also have more data, than for classifiers trained on age-based splits. For sex-based splits, the AUROC does not differ significantly across different data splits. For the age-based splits, the only significant difference across all split combinations concerns the 50/50 and 100/0 split, where the AUROC in the former is higher. A reason for this finding is likely that FL generally performs better for equally distributed data, instead of highly imbalanced data distributed across institutions [1, 5].

In both cases, sex-based and age-based splits, corresponding institutions (e.g., NIH-1 and NIH-2) do not differ significantly in their SV when we look at the 'as is' and 50/50 split. This result is expected, as these institutions use data following the same distribution. Regardless of the data split, in all cases, every institution contributes positively toward the AUROC of the predictive performance of the classifier on the test dataset. Institutions holding NIH subsets have lower SVs than those holding CXP or CXR subsets. For the CXP and CXR datasets, the highest mean SVs are for institutions with 100% female patient scans, respectively 100% young patient scans (CXP with 5.932% and 5.764%, whereas CXR with 5.985% and 6.314%). For the NIH dataset, the highest mean SVs are for the male 100/0 or the younger patient-dominated 75/25 age-based split.

---

[1] https://github.com/kpandl/Reward-System-for-Trustworthy-Medical-Federated-Learning

For the sex-based splits and the NIH clients, the male-dominated clients in the 75/25 and 100/0 splits contributed significantly more than the female-dominated clients. For the CXP and CXR datasets, there is no significant difference. However, in the age-based splits, younger-dominated clients contribute significantly more than older-dominated ones for the 100/0 case of the CXP clients and the 100/0 and 75/25 cases of the CXR clients.

Table 2: Mean SVs and 95% confidence interval per data split and client for the contributions toward the overall predictive performance. For the 'as is' and 50/50 splits, institutions with subsets from the same dataset (e.g., NIH-1, NIH-2) have the same data distribution. For the 75/25 and 100/0 splits, the 1st institution (e.g., NIH-1) has predominantly female, respectively young patient scans, whereas the 2nd institution (e.g., NIH-2) has predominantly male, respectively old patient scans. The sum of the SVs plus .5 (performance of a random classifier) equals the 'Total AUROC' column. The bold marked values represent the highest value of the two corresponding clients in any sex-based or age-based split.

| Split | | NIH-1 [%] | NIH-2 [%] | CXP-1 [%] | CXP-2 [%] | CXR-1 [%] | CXR-2 [%] | Total AUROC [%] |
|---|---|---|---|---|---|---|---|---|
| Sex | As is | 3.898±0.446 | 4.370±0.263 | 5.867±0.148 | 5.758±0.141 | 5.908±0.161 | 5.842±0.129 | 81.643±0.052 |
| | 50/50 | 3.935±0.558 | 4.413±0.289 | 5.905±0.174 | 5.780±0.151 | 5.775±0.220 | 5.818±0.163 | 81.626±0.046 |
| | 75/25 | 3.771±0.503 | 4.463±0.228 | 5.831±0.168 | 5.777±0.128 | 5.939±0.183 | 5.866±0.177 | 81.647±0.056 |
| | 100/0 | 3.307±0.421 | **4.771±0.212** | **5.932±0.148** | 5.790±0.138 | **5.985±0.189** | 5.837±0.168 | 81.622±0.049 |
| Age | As is | 3.842±0.538 | 3.941±0.266 | 5.676±0.218 | 5.645±0.155 | 6.036±0.120 | 6.035±0.084 | 81.175±0.057 |
| | 50/50 | 3.852±0.529 | 3.834±0.263 | 5.746±0.115 | 5.670±0.118 | 6.047±0.095 | 6.049±0.084 | 81.198±0.044 |
| | 75/25 | **3.884±0.519** | 3.829±0.242 | 5.704±0.130 | 5.667±0.116 | 6.170±0.085 | 5.930±0.092 | 81.184±0.047 |
| | 100/0 | 3.819±0.363 | 3.835±0.203 | **5.764±0.112** | 5.549±0.092 | **6.314±0.083** | 5.858±0.071 | 81.139±0.047 |

### 4.2 Contributions toward bias

The results of the contribution quantification toward bias are shown in Figure 1. The overall bias is positive for all sex-based splits and age-based splits, which is favorable for female and younger patients. Thereby, the age-related bias is much larger than the sex-related bias. Across the four different age-based splits, there is no statistically significant difference in the bias. Across the sex-based splits, the 'as is' split has a statistically significantly lower bias than the 50/50 and 75/25 splits, and the 100/0 split has a statistically significantly lower bias than the 75/25 split.

For the sex-based splits, the 'as is' NIH and CXR clients introduce a bias favoring male patients, whereas the mean contribution of the CXP clients introduce a bias favoring female patients. A general trend indicates that the higher the share of one group, the larger the contribution for a bias favoring that group. In the 100/0 split, all female patient clients introduce a bias favoring female patients, and all male patient clients introduce a bias favoring male patients. For the age-based 'as is' split, all clients introduce a bias favoring younger patients, irrespective of the data split. Overall, increasing the share of one subgroup in the institution's dataset increases the bias contribution of that institution favoring this subgroup. These results indicate that the SV can also quantify contributions toward bias in FL, which highly depend on the data distribution of the institutions. Institutions in a FL consortium may have different signs of their bias contributions, which can help to balance out and reduce the overall bias.

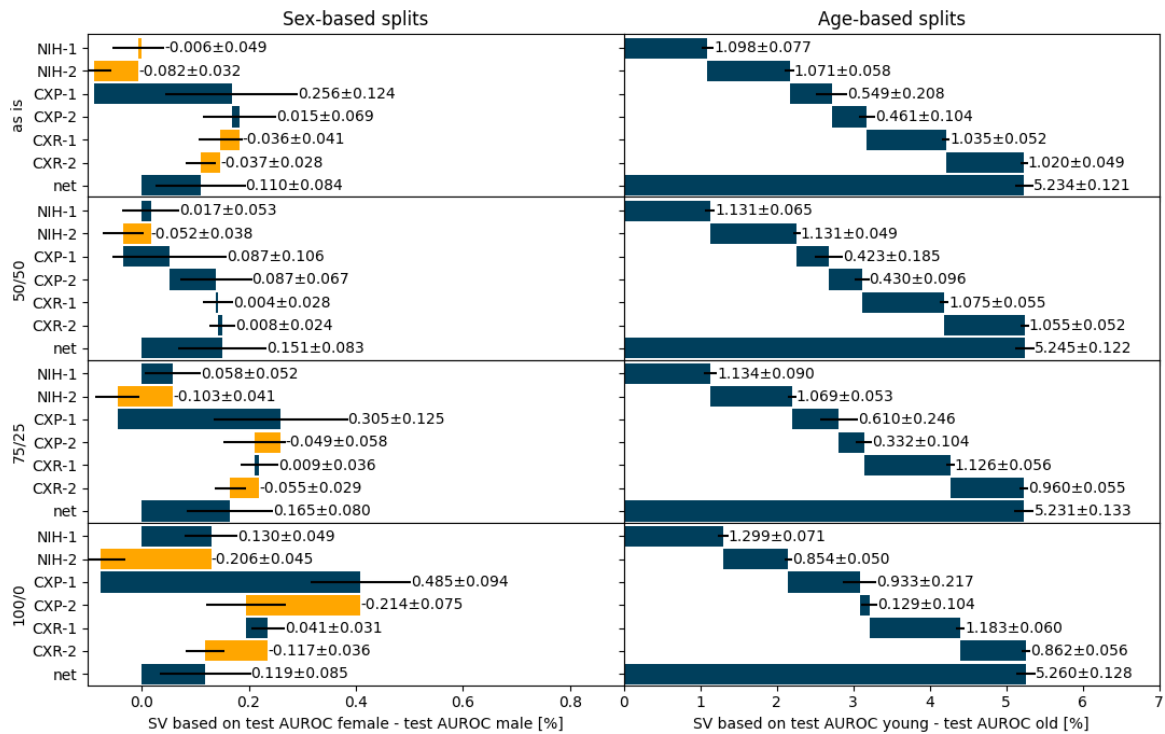

Figure 1: Mean SVs and 95% confidence intervals of the institutions of all 8 splits for contributions toward sex bias or age bias. Positive means are colored in dark blue, negative means in orange.

## 5 DESIGN OF REWARD SYSTEMS

### 5.1 Rewards for contributions toward predictive performance

Based on the results of the contributions on predictive performance in Table 2, we now aim to fairly distribute rewards amongst institutions. For this, we consider reward systems with a fixed budget defined upfront of the FL process. A coalition of institutions can obtain such a budget, for example, from an external funding agency, or from deposits of each participating institution. Alternatively, institutions can continuously share capital flows over an extended period of time, particularly when revenues from deploying the FL model occur later [8].

For distributing rewards toward predictive performance, we assume a reward pool $P$ of 60 monetary units (MU) for the 6 institutions. In case it originates from equal deposits of each institution, each institution would pay 10 MUs into the reward pool. Following the definition of the SV in II-C, the utility $U(D)$ of the entire coalition $D$ consisting of the institutions $z_i$ represents the gain in AUROC of the coalition compared to a random classifier with an AUROC of 0.5 and is equal to the sum of the SVs, as denoted in equation (2). Its value is at most 0.5, which means a perfect classifier has an AUROC of 1.

In line with the definition of the SV, each coalition of institutions gets rewarded in proportion to its success of contributing toward a high AUROC in an FL model. Therefore, we only want to reward the actual utility of a coalition in improving the classifier compared to a random classifier. For that, we introduce the size of the actually distributed reward pool $P_{dist}$ in equation (3).

$$P_{dist} = P \cdot \frac{U(D)}{0.5} \tag{3}$$

The remaining reward pool of size $P - P_{dist}$ can either be distributed back equally to the institutions or, in case it originates from an external source, be distributed back to the external source. We distribute the rewards $R(z_i)$ of an institution $z_i$ according to equation (4) in proportion to the SVs $\varphi(z_i)$.

$$R(z_i) = \frac{\varphi(z_i)}{U(D)} \cdot P_{dist} = \frac{\varphi(z_i)}{0.5} \cdot P \tag{4}$$

If the reward pool $P$ of 60 MUs comes from the institutions themselves, the profit per institution can be calculated based on equation (5).

$$G(z_i) = R(z_i) - \frac{P_{dist}}{|D|} \tag{5}$$

In some cases, it may also be desirable to distribute an entire reward pool independent of the utility of a coalition, which has been proposed by prior research [27]. Thereby, the reward distribution formula simplifies to equation (6). However, this comes with the risk that external agencies or institutions themselves pay potentially large amounts for an FL model, which may not be much better than a random classifier.

$$R(z_i) = \frac{\varphi(z_i)}{U(D)} \cdot P \tag{6}$$

The results of the reward distribution based on equation (4) are shown in Figure 2 in the blue bars. In the sex-based 'as is' split, the mean individual reward regarding predictive performance for the NIH institutions is 6.976 MUs, and 7.051 MUs for the CXR subset institutions. It is significantly lower for the CXP institutions with 4.961 MUs.

In the sex-based 100/0 split, the mean reward is 7.119 MUs for the NIH institution with female patient scans and 6.948 MUs for the NIH institution with male patient scans. For the CXP institutions, it is 3.968 MUs respectively 5.725 MUs, meaning higher for the client with male patient scans. For the CXR institutions, it is 7.182 and 7.004 MUs.

In the age-based 100/0 split, a similar pattern emerges where the mean rewards for younger patient institutions are higher than the mean rewards for older patient institutions (6.917 MUs and 6.659 MUs for NIH, 4.583 MUs and 4.602 MUs for CXP, 7.577 MUs and 7.030 MUs for CXR). Similar to the sex-based 'as is' split, both, sex-based and age-based 100/0 splits have significantly smaller the mean rewards for the CXP clients compared to the mean rewards of their NIH or CXR counterparts.

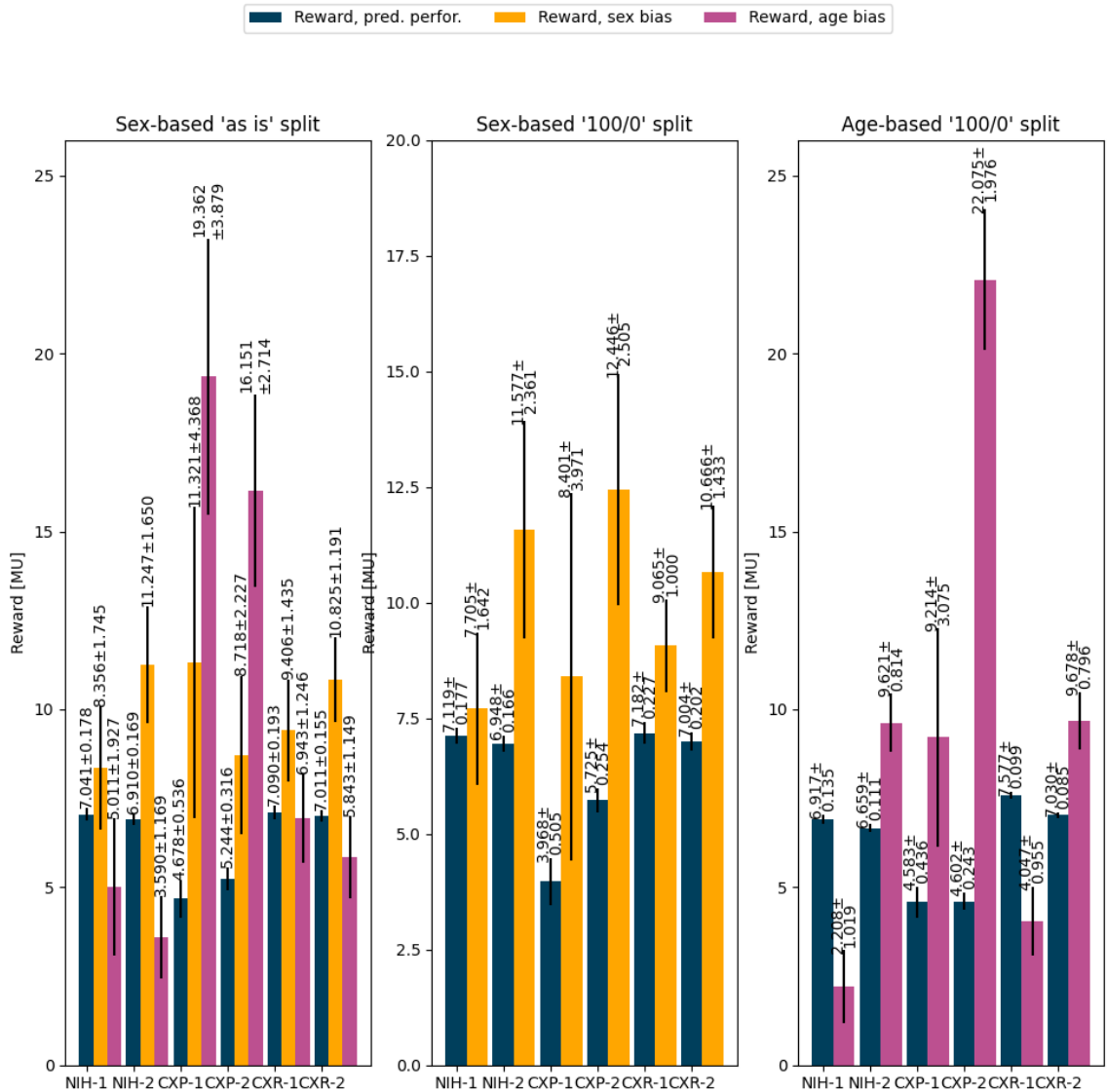

Figure 2: results on the rewards with regard to predictive performance, sex bias, and age bias for the sex-based 'as is' split. For the sex-based and age-based 100/0 splits, we computed results for the rewards with regard to predictive performance, and sex bias respectively, age bias.

### 5.2 Rewards for contributions toward bias

Since a low model bias is a goal for trustworthy ML in healthcare [4, 7], we aim to incentivize contributions toward a reduction in the absolute bias. In contrast to predictive performance, small or potentially negative-valued contributions can be more desirable than large contributions and should then be incentivized with a higher reward.

Here, the utility of the coalition $U(D)$ is the bias and therefore, equal to the sum of the SVs toward the bias. It can hypothetically range from -1 to 1, as it is the difference between two AUROC scores, each in the range from 0 to 1. Based on the idea that a low absolute bias is desired, we introduce the size of the actually distributed reward pool $P_{dist}$, as denoted in equation (7).

$$P_{dist} = P \cdot (1 - |U(D)|) \tag{7}$$

Like the rewards for predictive performance contributions, the remaining reward pool of size $P - P_{dist}$ can either be distributed back equally to the institutions, or, in case it originates from an external source, be distributed back to the external source.

In case the bias is not zero and institutions have different SVs, we first consider the sign of the bias $U(D)$ and identify the index $w$ of the institution with the highest contribution toward the absolute bias, as described in equation (8).

$$w = \underset{i}{\operatorname{argmax}} \operatorname{sgn}(U(D)) \cdot \varphi(z_i) \tag{8}$$

In a next step, we calculate the difference $\Delta(z_i)$ between the bias contribution of institution $z_w$ and the bias contribution of institution $z_i$ for all institutions.

$$\Delta(z_i) = \varphi(z_w) - \varphi(z_i) \tag{9}$$

One way to allocate the entire distribution reward pool $P_{dist}$ is by allocating rewards based on $\Delta(z_i)$ in proportion to the sum over all $\Delta(z_i)$, as denoted in equation (10).

$$R(z_i) = \frac{\Delta(z_i)}{\sum_{z_i \in D} \Delta(z_i)} \cdot P_{dist} = \frac{\Delta(z_i)}{|D| * \varphi(z_w) - U(D)} \cdot P_{dist} \tag{10}$$

Like before, if the reward pool comes from the institutions themselves, the profit of an institution is then calculated by subtracting the reward from the payment of an institution, cf. equation (11).

$$G(z_i) = R(z_i) - \frac{P_{dist}}{|D|} \tag{11}$$

In case the bias is zero or SVs among institutions are the same, all bias contributions balance, and a natural approach would be to distribute rewards equally among institutions.

For the sex-based 'as is' split, we compute the bias rewards with regard to patient sex and age (cf. Table 2 for the SVs). The results are shown in Figure 2. Based on equation (10), incentivizing contributions on bias regarding patient sex results in a mean reward of 9.802 MUs for the NIH clients, a mean reward of 11.284 MUs for the CXP clients, and a mean individual reward of 10.112 MUs for the CXR clients. When incentivizing contributions toward a bias in patient age, it results in a mean individual reward of 4.301 MUs for the NIH clients, a mean individual reward of 13.736 MUs for the CXP clients, and a mean individual reward of 6.393 MUs for the CXR clients.

In the sex-based 100/0 split, the mean rewards are 7.705 MUs for the NIH client with female patients and 11.577 MUs for the NIH client with male patients. For the other institutions, the pattern holds that female patient institutions are rewarded less than their male patient counterpart institutions (8.401 MUs to 12.446 MUs for CXP, 9.065 MUs to 10.666 MUs for CXR), while the overall bias of the model favors female patients.

Similarly, in the age-based 100/0 split, young patient institutions receive significantly less rewards than their counterpart old patient institutions (2.208 MUs and 9.621 MUs for NIH, 9.214 MUs and 22.075 MUs for CXP, and 4.047 MUs and 9.678 MUs for CXR).

### 5.3 Combined reward scheme

A combined reward scheme aims to incentivize contributions toward both, a high predictive performance, and a low bias. Institutions have different SVs toward these different metrics, which can be computed at the same time. We propose a setup with multiple reward pools to enable a combined reward scheme. Institutions thereby need to agree on the relative importance of objectives.

We study a combined reward scheme based on the sex-based 'as is' split, with three reward pools for predictive performance, a low bias toward patient sex, and a low bias toward patient age. The three objectives each have a reward pool with 60 MUs. The total reward of an institution then computes as a combination of the three reward pools, as denoted in equation (12).

$$R(z_i) = R_{pred.perf.}(z_i) + R_{sex\ bias}(z_i) + R_{age\ bias}(z_i) \tag{12}$$

The results compose from the individual reward pools shown in Figure 2. Thereby, the NIH institutions individually receive mean rewards of 21.078 MUs accumulated over all 3 rewards, the CXP institutions of 24.851 MUs, and the CXR institutions of 23.559 MUs. While the CXP institutions receive the lowest rewards for contributions toward predictive performance, their contributions toward both, a reduction in absolute sex bias and a reduction in absolute age bias strongly increase their total rewards. As a result, they receive higher rewards than the other institutions.

### 5.4 Label flip experiment and incentives in practice

The results of the label flip experiment are illustrated in Figure 3, showing the rewards for the clients depending on their share of flipped labels. Generally, clients with flipped labels receive lower rewards, and this effect increases with an increasing ratio of flipped labels. For example, the mean reward changes from 6.910 MUs, 5.244 MUs and 7.011 MUs for no flipped labels for the respective clients to 6.072 MUs, 3.030 MUs, and 5.851 MUs for clients with flipped labels. Interestingly, the mean rewards also increase for the institutions with unchanged labels to 7.335 MUs, 7.891 MUs, and 7.056 MUs. Thereby, the effect is much larger for the NIH clients than for the other clients. The results indicate that institutions with good datasets and no flipped labels potentially equalize some of the adversarial effect by institutions with a higher share of wrong labels. The results also show that clients with non-flipped labels can expect to receive higher rewards than clients with flipped labels. This suggests that a reward system for predictive performance incentivizes institutions to invest in the label quality of their ML datasets.

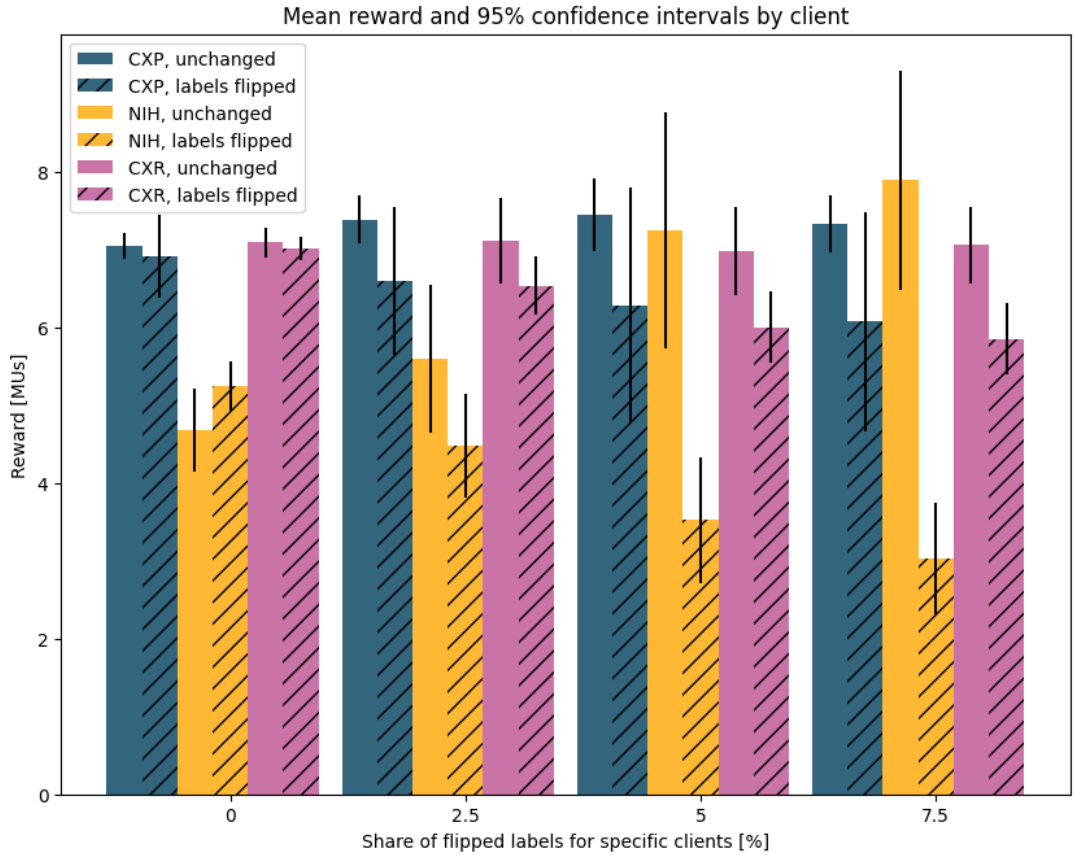

Figure 3: results on the rewards regarding predictive performance in the label flip experiment. The mean and confidence intervals are based out of 12 experiments.

**5.5 Scalability**

The scalability of the reward systems depends on the scalability of approximating the SVs. The SV complexity is generally of exponential form as $2^{n-1}$ coalitions need to be trained and tested for $n$ clients. As such, computing canonical SVs is highly expensive. Using an approximated SV estimation method, we only need to train the largest coalition once, and test all possible coalitions. However, the testing complexity is still exponential. The testing of a coalition is relatively quick, as it involves accumulating the outputs of the different LR classifiers [21] from the participating institutions. For example, to test the LR classifiers for all combinations of six FL clients, evaluations of 63 coalitions were necessary with a total compute time of 4.56 s, or mean time of 72.38 ms per coalition. Generally, cross-silo FL consortia, have up to 100 institutions [5]. In healthcare specifically, one of the larger FL consortia for clinical outcome prediction of COVID-19 patients had 20 participating institutions [37]. As a result, the expected computational time for the SV approximation would be around 21 hours, which is still feasible. Thus, the approach scales reasonably well for established consortia. Furthermore, there is potential room for performance improvement in the implementation (e.g., parallelization, using compiled programming languages instead of Python, or faster storage access), and the authors of the SV approximation method show that even 30 participating institutions could be feasible [21]. Nevertheless, novel SV estimation techniques may be required for even larger consortia in the future.

## 6 DISCUSSION

### 6.1 Principal findings

Our FL experiments with six institutions sampled from the NIH, CXP, and CXR datasets suggest that there is an AUROC bias favorable for female patients and young patients. Thereby, the bias for young patients is much larger than the bias for female patients. Other extant research on bias in a combination of these datasets on centralized ML identifies a bias for four out of seven disease labels favoring male patients, and five out of seven disease labels favoring older patients [18]. However, these results are not directly comparable with our results, as their focus is on a sensitivity disparity, the evaluation metric is a label count, and centralized ML instead of FL [18].

Considering this existing bias, the developed reward systems also provide some interesting insights. We find that in many cases, institutions with predominantly favored patient subgroups (female patients, young patients) have higher contributions toward predictive performance, and as a result, also receive more rewards for their contributions. On the other hand, counterparty institutions with predominantly disfavored patient subgroups (male patients, old patients) often have lower contributions toward predictive performance and receive less rewards as well. These results indicate that a model bias affecting patients can also convey to an institutional level: institutions with advantaged patients receive higher rewards than institutions with disadvantaged patients. For reward systems in practice, this can economically handicap institutions of disadvantaged patients, and thus, reinforce back to the patients (e.g., through worsened equipment in the financially disadvantaged institution). We address this issue by developing a reward system that incentivizes contributions toward a reduction of the absolute bias. Analogously, the rewards are allocated in proportion to the utility of the coalition. We evaluate it for contributions toward sex bias and age bias. We then combine the reward system incentivizing predictive performance, and the ones incentivizing a reduction in bias. The results show that a combined reward system can incentivize trustworthy AI in terms of high predictive performance and low bias. Noteworthy, a reward system can also incentivize institutions to invest in their dataset label quality. Institutions with low error-rate datasets can also receive higher rewards when other institutions have a higher rate of label errors in their datasets.

Regarding the contribution quantification, our results suggest that SV approximations can be successfully used to also quantify contributions toward bias and scale well for even larger contemporary FL consortium sizes. The results show that institutions with a higher share of scans sampled from one subgroup introduce a stronger bias favoring this subgroup, compared with institutions sampling the other subgroup from the same dataset. In this regard, FL behaves similar to centralized ML, where prior research found this tendency with regard to sex bias [19]. Thereby, we also saw that the partitioning of scans only had a significant impact for certain partitions in terms of age but not regarding sex, and even then, the impact is relatively small compared to the bias. This result is interesting in the light of prior concerns that a heterogeneous data partitioning leads to a strongly increased bias in FL [6].

### 6.2 Implications for research and practice

Our work has several implications for research and practice. For research, our results show the necessity that FL institutions require means to better understand and control the bias they introduce to a FL consortium [5, 38, 39]. This may help institutions in receiving higher rewards, and ultimately, reduce the bias of the global FL model. Furthermore, our results ask for more economic research on reward systems for medical FL. This includes the question of the importance of bias for different subgroups in healthcare [18], and the relative importance of a low absolute bias compared to a high predictive performance in FL, as this would influence the distribution of different reward pools. Further research may also deploy reward systems in healthcare practice and investigate the practical benefits and challenges. Thereby, it can also aim to find out more about the readiness of institutions in adopting reward systems. Finally, a further interesting avenue for research is the design of reward systems in decentralized swarm learning scenarios. In doing so, a blockchain-based system orchestrates the FL process with promising applications for healthcare [16, 40]. Future research may aim to further integrate reward systems into automated, smart contract-based data marketplaces [16].

For practice, our findings suggest that FL models trained on real-world datasets often come with an inherent bias. Practitioners need to cautiously study potential implications when designing and deploying reward systems. Otherwise, reward systems can convey a model bias to an institutional level, and potentially amplify devastating effects of bias in healthcare. If reward systems properly account for bias contributions, our results suggest that reward systems can provide incentives for trustworthy AI in healthcare and that these are ready for deployment and evaluation in practice.

### 6.3 Limitations and future research

The limitations of our study are as follows. First, FL institutions may make malicious contributions intentionally or unintentionally, which may result in a negative SV contribution for their predictive performance and demands for an extension of our reward allocation scheme for predictive performance in future research. Similarly, further design possibilities exist for reward systems that incentivize contributions toward bias, besides the one we developed. One alternative way would be to proportionally scale rewards such that an institution with zero bias contribution has a profit of zero. However, this may not allocate the entire distribution reward pool $P_{dist}$. Therefore, we leave the development and comparison of further reward systems for bias contributions in FL to future research. Second, our developed reward systems reward retroactively, only after the training process is finished. In case institutions fund the reward pools, this can come with economic risk for institutions: for example, one institution may have different scans than all other institutions because it has a different scanner type. This could lead to disadvantages in the reward system. This is especially relevant to bias, where it may be difficult to estimate the sign of the final FL model bias before the training. As a result, it is difficult for institutions to estimate whether they will reduce the overall bias, or amplify it, which has a strong implication for their received rewards. This limitation asks for future research tackling estimation of model bias in FL scenarios, and the design of reward systems where institutions can dynamically decide to remain or leave an FL consortium at an early stage.

Third, due to the design of the available datasets, we only analyze a limited set of diseases and bias references. Future research may analyze the findings on further diseases, other bias references (e.g., ethnicity, insurance or socioeconomic status, preconditions) and on medical datasets other than chest X-ray datasets.

Last, we use the same dataset size for all institutions to distill the influence of data partitioning on the bias. Future research may analyze the impact of different institution dataset sizes in a federated weighted averaging scenario on the model bias, which is an open question [6], as well as the impact on FL reward systems that incentivize bias contributions.

## 7 CONCLUSION

In this research, we first measure contributions of FL institutions toward both, predictive performance, and model bias. Based on the SV approximations, we then develop reward systems that incentivizes contributions toward trustworthy AI by incentivizing outcomes with high predictive performance and low absolute bias. We evaluate the reward systems and show that they can effectively incentivize a high predictive performance and a low bias. Furthermore, they can incentivize FL institutions to have high-quality datasets with reduced label errors. The approach scales well to even larger contemporary medical FL consortium sizes.

We contribute to the extant body of knowledge with our research in three ways. First, we analyze the model bias in a medical FL scenario, answering prior calls for research in this area. Thereby, we identify a small influence of the partitioning of chest X-ray scans across different institutions on the model bias in FL in some instances. Second, we demonstrate that SV approximations can not only quantify contributions toward predictive performance, but also bias in medical FL. Third, we develop reward systems that incentivize contributions of FL institutions toward predictive performance and model bias. Thereby, we answer previous calls for research on reward systems for FL and incentives for trustworthy AI.

## ACKNOWLEDGMENTS

The authors acknowledge support by the state of Baden-Württemberg through bwHPC and the German Research Foundation (DFG) through grant INST 35/1134-1 FUGG.

## REFERENCES

[1] N. Rieke *et al.*, "The future of digital health with federated learning," *NPJ digital medicine,* vol. 3, no. 1, pp. 1-7, 2020.
[2] D. Ng, X. Lan, M. M.-S. Yao, W. P. Chan, and M. Feng, "Federated learning: a collaborative effort to achieve better medical imaging models for individual sites that have small labelled datasets," *Quantitative Imaging in Medicine and Surgery,* vol. 11, no. 2, p. 852, 2021.
[3] Y. Zhao, M. Li, L. Lai, N. Suda, D. Civin, and V. Chandra, "Federated learning with non-iid data," *arXiv preprint arXiv:1806.00582,* 2018.
[4] S. Thiebes, S. Lins, and A. Sunyaev, "Trustworthy artificial intelligence," *Electronic Markets,* vol. 31, no. 2, pp. 447-464, 2021.
[5] P. Kairouz *et al.*, "Advances and open problems in federated learning," *Foundations and Trends® in Machine Learning,* vol. 14, no. 1–2, pp. 1-210, 2021.
[6] A. Abay, E. Chuba, Y. Zhou, N. Baracaldo, and H. Ludwig, "Addressing Unique Fairness Obstacles within Federated Learning," *AAAI 2021 workshop,* 2021.
[7] J. Wiens *et al.*, "Do no harm: a roadmap for responsible machine learning for health care," *Nature medicine,* vol. 25, no. 9, pp. 1337-1340, 2019.
[8] S. Yang, F. Wu, S. Tang, X. Gao, B. Yang, and G. Chen, "On designing data quality-aware truth estimation and surplus sharing method for mobile crowdsensing," *IEEE Journal on Selected Areas in Communications,* vol. 35, no. 4, pp. 832-847, 2017.
[9] Z. Zhou, L. Chu, C. Liu, L. Wang, J. Pei, and Y. Zhang, "Towards Fair Federated Learning," in *Proceedings of the 27th ACM SIGKDD Conference on Knowledge Discovery & Data Mining*, 2021, pp. 4100-4101.
[10] N. Evangelatos *et al.*, "Digital transformation and governance innovation for public biobanks and free/Libre open source software using a blockchain technology," *OMICS: A Journal of Integrative Biology,* vol. 24, no. 5, pp. 278-285, 2020.

[11]   T. Song, Y. Tong, and S. Wei, "Profit allocation for federated learning," in *2019 IEEE International Conference on Big Data (Big Data)*, 2019: IEEE, pp. 2577-2586.
[12]   T. Wang, J. Rausch, C. Zhang, R. Jia, and D. Song, "A principled approach to data valuation for federated learning," in *Federated Learning*: Springer, 2020, pp. 153-167.
[13]   Z. Liu, Y. Chen, H. Yu, Y. Liu, and L. Cui, "GTG-Shapley: Efficient and Accurate Participant Contribution Evaluation in Federated Learning," *arXiv preprint arXiv:2109.02053,* 2021.
[14]   A. Ghorbani and J. Zou, "Data shapley: Equitable valuation of data for machine learning," in *International Conference on Machine Learning*, 2019: PMLR, pp. 2242-2251.
[15]   K. D. Pandl, F. Feiland, S. Thiebes, and A. Sunyaev, "Trustworthy machine learning for health care: scalable data valuation with the shapley value," in *Proceedings of the Conference on Health, Inference, and Learning*, 2021, pp. 47-57.
[16]   K. D. Pandl, S. Thiebes, M. Schmidt-Kraepelin, and A. Sunyaev, "On the convergence of artificial intelligence and distributed ledger technology: A scoping review and future research agenda," *IEEE access,* vol. 8, pp. 57075-57095, 2020.
[17]   S. Feuerriegel, M. Dolata, and G. Schwabe, "Fair AI," *Business & information systems engineering,* vol. 62, no. 4, pp. 379-384, 2020.
[18]   L. Seyyed-Kalantari, G. Liu, M. McDermott, I. Y. Chen, and M. Ghassemi, "CheXclusion: Fairness gaps in deep chest X-ray classifiers," in *BIOCOMPUTING 2021: Proceedings of the Pacific Symposium*, 2020: World Scientific, pp. 232-243.
[19]   A. J. Larrazabal, N. Nieto, V. Peterson, D. H. Milone, and E. Ferrante, "Gender imbalance in medical imaging datasets produces biased classifiers for computer-aided diagnosis," *Proceedings of the National Academy of Sciences,* vol. 117, no. 23, pp. 12592-12594, 2020.
[20]   A. Xu *et al.*, "Closing the Generalization Gap of Cross-silo Federated Medical Image Segmentation," *arXiv preprint arXiv:2203.10144,* 2022.
[21]   S. Kumar *et al.*, "Towards More Efficient Data Valuation in Healthcare Federated Learning Using Ensembling," in *Distributed, Collaborative, and Federated Learning, and Affordable AI and Healthcare for Resource Diverse Global Health: Third MICCAI Workshop, DeCaF 2022, and Second MICCAI Workshop, FAIR 2022, Held in Conjunction with MICCAI 2022, Singapore, September 18 and 22, 2022, Proceedings*, 2022: Springer, pp. 119-129.
[22]   Y. Wu, J. Zhu, Y. Yuan, and D. Cheng, "Profit Sharing Method for Participants in Cloud Energy Storage," in *2020 12th IEEE PES Asia-Pacific Power and Energy Engineering Conference (APPEEC)*, 2020: IEEE, pp. 1-5.
[23]   N. Hynes, D. Dao, D. Yan, R. Cheng, and D. Song, "A demonstration of sterling: a privacy-preserving data marketplace," *Proceedings of the VLDB Endowment,* vol. 11, no. 12, pp. 2086-2089, 2018.
[24]   A. Agarwal, M. Dahleh, and T. Sarkar, "A marketplace for data: An algorithmic solution," in *Proceedings of the 2019 ACM Conference on Economics and Computation*, 2019, pp. 701-726.
[25]   J. Zhang, Y. Wu, and R. Pan, "Incentive mechanism for horizontal federated learning based on reputation and reverse auction," in *Proceedings of the Web Conference 2021*, 2021, pp. 947-956.
[26]   H. Yin, A. Mallya, A. Vahdat, J. M. Alvarez, J. Kautz, and P. Molchanov, "See through gradients: Image batch recovery via gradinversion," in *Proceedings of the IEEE/CVF Conference on Computer Vision and Pattern Recognition*, 2021, pp. 16337-16346.
[27]   Y. Liu, Z. Ai, S. Sun, S. Zhang, Z. Liu, and H. Yu, "Fedcoin: A peer-to-peer payment system for federated learning," in *Federated Learning*: Springer, 2020, pp. 125-138.
[28]   L. Gao, L. Li, Y. Chen, W. Zheng, C. Xu, and M. Xu, "FIFL: A Fair Incentive Mechanism for Federated Learning," in *50th International Conference on Parallel Processing*, 2021, pp. 1-10.
[29]   X. Wang, Y. Peng, L. Lu, Z. Lu, M. Bagheri, and R. M. Summers, "Chestx-ray8: Hospital-scale chest x-ray database and benchmarks on weakly-supervised classification and localization of common thorax diseases," in *Proceedings of the IEEE conference on computer vision and pattern recognition*, 2017, pp. 2097-2106.
[30]   J. Irvin *et al.*, "Chexpert: A large chest radiograph dataset with uncertainty labels and expert comparison," *arXiv preprint arXiv:1901.07031,* 2019.
[31]   A. E. Johnson *et al.*, "MIMIC-CXR, a de-identified publicly available database of chest radiographs with free-text reports," *Scientific data,* vol. 6, no. 1, pp. 1-8, 2019.

[32] G. Huang, Z. Liu, L. Van Der Maaten, and K. Q. Weinberger, "Densely connected convolutional networks," in *Proceedings of the IEEE conference on computer vision and pattern recognition*, 2017, pp. 4700-4708.
[33] P. Rajpurkar *et al.*, "Chexnet: Radiologist-level pneumonia detection on chest x-rays with deep learning," *arXiv preprint arXiv:1711.05225,* 2017.
[34] B. McMahan, E. Moore, D. Ramage, S. Hampson, and B. A. y Arcas, "Communication-efficient learning of deep networks from decentralized data," in *Artificial intelligence and statistics*, 2017: PMLR, pp. 1273-1282.
[35] T. Rädsch, S. Eckhardt, F. Leiser, K. D. Pandl, S. Thiebes, and A. Sunyaev, "What your radiologist might be missing: using machine learning to identify mislabeled instances of X-ray images," 2021.
[36] F. Leiser *et al.*, "Understanding the Role of Expert Intuition in Medical Image Annotation: A Cognitive Task Analysis Approach," in *Proceedings of the 56th Hawaii International Conference on System Sciences (HICSS 2023)*, 2023.
[37] I. Dayan *et al.*, "Federated learning for predicting clinical outcomes in patients with COVID-19," *Nature medicine,* vol. 27, no. 10, pp. 1735-1743, 2021.
[38] W. Du, D. Xu, X. Wu, and H. Tong, "Fairness-aware agnostic federated learning," in *Proceedings of the 2021 SIAM International Conference on Data Mining (SDM)*, 2021: SIAM, pp. 181-189.
[39] Y. H. Ezzeldin, S. Yan, C. He, E. Ferrara, and S. Avestimehr, "Fairfed: Enabling group fairness in federated learning," *arXiv preprint arXiv:2110.00857,* 2021.
[40] S. Warnat-Herresthal *et al.*, "Swarm learning for decentralized and confidential clinical machine learning," *Nature,* vol. 594, no. 7862, pp. 265-270, 2021.